\documentclass[11pt]{article}

\usepackage[T1]{fontenc}
\usepackage{lmodern}
\usepackage[margin=1in]{geometry}
\usepackage[english]{babel}
\usepackage{amsmath,amssymb,mathtools}
\usepackage{array,booktabs,tabularx}
\usepackage{graphicx}
\usepackage{float}
\usepackage{xcolor}
\usepackage{microtype}
\usepackage{enumitem}
\usepackage{placeins}
\usepackage{subcaption}
\usepackage{listings}
\usepackage[numbers,sort&compress]{natbib}
\usepackage{hyperref}
\usepackage[nameinlink,noabbrev]{cleveref}
\usepackage{graphcore_report}

\title{Fast Polynomial Transcendentals for LLMs}
\reportsubtitle{Target-Precision Programs under Asymmetric GPU Scaling}
\reportfocus{Low-degree polynomial transcendentals for attention, dense activations, and routed experts.}
\reportfooter{Fast polynomial transcendentals}
\reportemail{Correspondence: \href{mailto:robert.stats.hu@gmail.com}{\nolinkurl{robert.stats.hu@gmail.com}}}
\author{Robert Hu}
\date{Sept 2026}
\hypersetup{
  colorlinks=true,
  linkcolor=GraphcoreInk,
  citecolor=GraphcoreCoralDark,
  urlcolor=GraphcoreCoralDark,
  pdftitle={Fast Polynomial Transcendentals for LLMs},
  pdfauthor={Robert Hu},
  pdfsubject={Target-precision polynomial transcendental functions under asymmetric GPU scaling},
  pdfkeywords={GPU kernels, polynomial approximation, transcendental functions, FlashAttention, mixture of experts}
}

\newcommand{\silu}{\operatorname{SiLU}}
\newcommand{\softcap}{\operatorname{softcap}}
\newcommand{\poly}{\mathrm{poly}}
\newcommand{\native}{\mathrm{native}}
\newcolumntype{Y}{>{\raggedright\arraybackslash}X}
\newcolumntype{C}{>{\centering\arraybackslash}X}
\newcommand{\reporttablefont}{\small}
\lstdefinelanguage{PTX}{
  alsoletter={.},
  morekeywords={.reg,.b32,cvt.rn.bf16x2.f32,mov.b32,
    set.ge.u32.bf16x2,min.bf16x2,max.bf16x2,lop3.b32,
    fma.rn.bf16x2,mul.rn.bf16x2},
  sensitive=true,
  morecomment=[l]{//}
}
\lstdefinestyle{reportcode}{
  basicstyle=\ttfamily\scriptsize,
  keywordstyle=\color{GraphcoreCoralDark}\bfseries,
  commentstyle=\color{GraphcoreMuted}\itshape,
  numbers=left,
  numberstyle=\tiny\color{GraphcoreMuted},
  numbersep=7pt,
  columns=fullflexible,
  keepspaces=true,
  breaklines=true,
  showstringspaces=false,
  frame=single,
  framesep=5pt,
  rulecolor=\color{GraphcoreLine},
  backgroundcolor=\color{GraphcorePaper},
  xleftmargin=1.8em,
  framexleftmargin=1.4em,
  captionpos=b,
  aboveskip=1em,
  belowskip=1em
}
\crefname{lstlisting}{listing}{listings}
\Crefname{lstlisting}{Listing}{Listings}
\begin{document}
\maketitle

\begin{abstract}
Graphics processing unit (GPU) generations scale matrix, special-function,
and memory pipelines at different rates, so kernel bottlenecks move as hardware
evolves. FlashAttention-4 exposed this imbalance inside attention on NVIDIA
Blackwell. We test whether short polynomial programs can accelerate other
special-function-unit (SFU) operations in large language models (LLMs). We
first compare native PyTorch evaluation with packed fused multiply--add
(FMA) programs in an isolated IEEE binary16 (FP16) sweep spanning L2-resident
and high-bandwidth-memory (HBM)-resident working sets. We then replace native
sigmoid, tanh, and sigmoid linear unit (SiLU) with degree-3 or degree-4
bfloat16 (BF16) programs in four GB200 integration tasks: dense SiLU,
tanh-softcapped attention, sigmoid attention, and routed-expert Swish-gated linear unit
(SwiGLU). The programs combine analytical symmetry, target-format rounding,
and packed arithmetic inside consuming kernels. The isolated paths improve by
1.19--2.19x in L2 and 1.00--1.70x in HBM. The dense-SiLU,
tanh-softcapped-attention, and routed-expert substitutions improve complete
training-step throughput by 2.7\%, 2.9\%, and 8.0\%, respectively. The
sigmoid-attention substitution improves complete-attention forward by 7.4\%
and the complete GPU step by 0.3\%. Same-checkpoint open-weight ablations and
one paired pre-training comparison per task extend the evaluation to model
behavior. At common horizons near 100 billion tokens, the final smoothed
training-loss differences (polynomial minus native) range from $-0.107$ to
$+0.079$ across the four tasks.
\end{abstract}


\section{Introduction}

Graphics processing unit (GPU) generations scale matrix,
special-function-unit (SFU), and memory pipelines at different rates.
FlashAttention-4 (FA4) provides a concrete example: on Blackwell, Tensor Core
throughput increased more quickly than the resources used for exponentials and
shared-memory traffic, exposing non-matrix work inside an optimized attention kernel
\cite{zadouri2026flashattention4}. FA4 recovered performance by redesigning the
algorithm and its instruction pipeline around that imbalance.

Large language models (LLMs) expose the same resource imbalance beyond softmax.
Dense multilayer perceptrons (MLPs), attention variants, and routed experts
interleave matrix multiplication with sigmoid, tanh, and sigmoid linear unit
(SiLU). The native forward paths studied here evaluate these transcendentals
through SFU-backed instruction sequences. Native backward paths either
reevaluate the required transcendental or apply an algebraic rule to a forward
result, depending on the operator. A short polynomial can instead use packed
bfloat16 (BF16) or IEEE binary16 (FP16) arithmetic on CUDA-core pipelines. We
trace each candidate from target-precision error through consuming-component
runtime and complete-step throughput, then measure same-checkpoint task
accuracy, word perplexity, prefill and decode throughput, and paired
pre-training loss.

We investigate the four integration tasks summarized in \cref{tab:cases}.

\begin{table}[H]
\centering
\scriptsize
\caption{The four principal interventions. D3 and D4 denote degree-3 and
degree-4 polynomials. B3 replaces sigmoid and its derivative; softmax
exponential is outside its scope.}
\label{tab:cases}
\begin{tabular*}{\textwidth}{@{\extracolsep{\fill}}p{0.06\textwidth}p{0.22\textwidth}p{0.27\textwidth}p{0.35\textwidth}@{}}
\toprule
\textbf{Case} & \textbf{Baseline} & \textbf{Polynomial intervention} &
\textbf{Question} \\
\midrule
B1 & Native SiLU in dense MLPs & Packed D3 SiLU &
Does packed SiLU increase dense-model training-step throughput? \\
B2 & Tanh-softcapped attention & D4 tanh inside FA4 &
Does packed tanh reduce complete softcapped-attention runtime? \\
B3 & Native sigmoid attention & Direct D3 sigmoid and D4 derivative &
Does polynomial sigmoid reduce complete sigmoid-attention runtime? \\
B4 & Native SiLU in routed experts & Fused D3 polynomial
Swish-gated linear unit (SwiGLU) &
Does packed SwiGLU increase routed-expert training-step throughput? \\
\bottomrule
\end{tabular*}
\end{table}

Throughout, ``native'' denotes the unchanged implementation of the same
transcendental for the same integration task.

This paper makes three contributions:
\begin{itemize}[leftmargin=1.4em]
    \item We construct target-precision programs by combining analytical range
    reduction, structurally constrained fitting, and replay of coefficient and
    intermediate rounding in deployed order. Same-form Sollya minimax fits
    provide coefficient controls.
    \item We lower the accepted programs to packed CUDA and inline Parallel
    Thread Execution (PTX) code in four consuming LLM components. B1, B2, and
    B4 improve complete-step throughput by 2.7\%, 2.9\%, and 8.0\%; B3 improves
    complete attention forward by 7.4\%. We release the fitting tools, packed
    kernels, portable integration and evaluation code, and supporting artifacts;
    repository and reproduction details appear in \cref{app:reproducibility}.
    \item We evaluate degree-4 substitutions in five open-weight models and
    compare native and polynomial pre-training trajectories for each case. At
    common horizons near 100 billion tokens, final smoothed training-loss
    differences are $+0.079$, $+0.003$, $-0.107$, and $+0.003$ for B1--B4.
\end{itemize}

\section{Background and Related Work}

\subsection{Asymmetric accelerator scaling}

Tensor Cores execute the matrix operations that dominate Transformer layers,
whereas CUDA-core and SFU pipelines execute scalar and packed arithmetic around
those matrices. The L2 cache and high-bandwidth memory (HBM) supply different
working-set regimes. Accelerating one resource changes which of the others is
visible in an integrated kernel.

FA4 quantified this effect on Blackwell: matrix throughput increased much more
quickly than the exponential units used by softmax, allowing non-matrix work to
occupy a substantial share of the attention schedule
\cite{zadouri2026flashattention4}. Blackwell Ultra subsequently doubled SFU
throughput relative to Blackwell for key attention instructions; NVIDIA
identifies the hardware exponential as \texttt{MUFU.EX2} in NVIDIA machine code
(SASS)
\cite{aubrey2025blackwellultra,nvidia2026blackwellultrasoftmax}.
NVIDIA reports that Rubin, the GPU in the Vera Rubin platform, provides up to
22~TB/s of HBM4 bandwidth, doubles Tensor Core throughput per clock relative
to Blackwell, and raises exponential throughput by 2x for 32-bit
floating-point (FP32) arithmetic and 4x for
BF16/FP16 \cite{alvarez2026rubinarchitecture}. These vendor-reported changes
illustrate why the active bottleneck can move between hardware generations;
all measurements in this paper use GB200.

\subsection{Target-precision numerical approximation}

Sollya provides high-precision function analysis, minimax construction,
coefficient rounding, and code-generation support \cite{sollya}. We use it as
the reference for unconstrained minimax fits. RLIBM instead constructs
polynomials around rounding intervals, demonstrating that output
representation and rounding behavior belong inside the approximation problem
\cite{lim2021rlibm32}. RLIBM targets correctly rounded libraries; our target is
a lower-latency program inside a specific BF16/FP16 neural-network kernel.

AutoNumerics-Zero searches directly over limited-precision programs and finds
approximations that trigger favorable compiler paths
\cite{real2023autonumerics}. Our search instead fixes analytical symmetry,
saturation, and a fused-multiply--add (FMA) budget before optimizing the
rounded program that executes.

K-TanH approximates tanh using low-precision integer operations, derives
sigmoid, Swish, and Gaussian error linear unit (GELU) variants, and reports
both CPU speed and Google Neural Machine Translation (GNMT) model convergence
\cite{kundu2019ktanh}. Tensor Processing Primitives use a
range-limited tanh polynomial evaluated with four Horner FMAs in a portable CPU
tensor library \cite{georganas2022tensor}. Our work targets packed BF16 using
CUDA and CuTe, NVIDIA's C++ CUDA template abstractions for hierarchical tensor
layouts, inside Blackwell attention and expert kernels rather than portable CPU
operators.

Non-uniform linear interpolation (NLI) selects interpolation cut points by
dynamic programming and applies the resulting piecewise-linear representation
to nonlinear operators including SiLU, root mean square layer normalization
(RMSNorm), and Softmax
\cite{yu2026nli}. NLI proposes a universal hardware unit; our programs remain
register-resident and use packed arithmetic on an unmodified GPU.

\subsection{Nonlinear operators in LLM components}

Low-Cost FlashAttention proposes fused exponential-and-multiplication hardware
\cite{alexandridis2025lowcostflashattention}; we instead change the software
program mapped to existing hardware. Sigmoid self-attention can match softmax
attention when normalized and stabilized appropriately, and FlashSigmoid
provides a memory-efficient implementation \cite{ramapuram2024sigmoidattention}.
B3 adopts that algorithmic setting and compares an SFU sigmoid with a packed
polynomial sigmoid.

Gemma 2 applies
\begin{equation}
    \softcap_{\alpha}(x)=\alpha\tanh(x/\alpha)
    \label{eq:softcap}
\end{equation}
to attention logits \cite{gemmateam2024gemma2}, giving B2 a direct open-weight
analogue. SwiGLU was established as a strong Transformer feed-forward choice
\cite{shazeer2020glu}, and DeepSeek-V3 makes routed SwiGLU experts a
representative production-scale mixture-of-experts (MoE) workload
\cite{deepseekai2024deepseekv3}.
These architectures motivate B1 and B4.

\section{Target-Precision Polynomial Programs}

Our optimization object is the complete low-precision program in BF16 or FP16:
analytical range reduction, polynomial coefficients, instruction order,
intermediate rounding, tail behavior, and the backward rule. A real-arithmetic
polynomial can change materially after its coefficients and Horner stages are
rounded. We therefore choose the mathematical form first, fit its residual
behavior, and replay the exact target-precision schedule before integrating it
into a kernel. We write D$k$ for a degree-$k$ polynomial.

\subsection{Analytical reduction}

Analytical identities reduce the fitted domain and remove coefficients for
known offsets, linear terms, and reflected values. Let
\begin{equation}
    \hat{x}=\operatorname{clip}(x,-L_c,L_c), \qquad t=|\hat{x}|.
    \label{eq:reduction}
\end{equation}
For sigmoid, the centered function
$g_{\sigma}(x)=\sigma(x)-\tfrac{1}{2}$ is odd because
$\sigma(-x)=1-\sigma(x)$. We therefore fit a polynomial $p_D$ only on the
positive half-domain and reconstruct
\begin{equation}
    \widetilde{\sigma}(x)
    =\frac{1}{2}+\operatorname{sign}(x)p_D(t),
    \qquad p_D(0)=0.
    \label{eq:sigmoid}
\end{equation}
The same construction applies to the odd function tanh. Sigmoid and tanh
derivatives are even, so their direct backward programs evaluate a polynomial
of $t$ without sign reconstruction. Saturating targets use a clamp or an exact
tail outside the active interval. This both prevents polynomial blow-up on
outliers and lets a short interior polynomial spend its error budget where the
function has useful curvature.

Although $s(x)=x\sigma(x)$ is neither odd nor even, its asymmetric part is
known exactly:
\begin{align}
    s(x)-s(-x)&=x, \\
    s(x)-\frac{x}{2}&=x\left(\sigma(x)-\frac{1}{2}\right).
    \label{eq:silu-decomposition}
\end{align}
The residual in the second line is even, because it is the product of two odd
functions. Consequently the forward program can be written as
\begin{equation}
    \widetilde{s}(x)=\frac{x}{2}+t\,p_D(t),
    \label{eq:silu}
\end{equation}
without constructing a full sigmoid tensor. Exact GELU, $x\Phi(x)$, has the
same decomposition since $\Phi(-x)=1-\Phi(x)$. These identities are exact;
the approximation is confined to one half-domain residual.

\begin{figure}[H]
\centering
\includegraphics[
  width=\textwidth,
  alt={Three panels showing symmetry reduction for sigmoid, SiLU, and the SiLU derivative; the positive half-domain is fitted and the negative half is reconstructed analytically.}
]{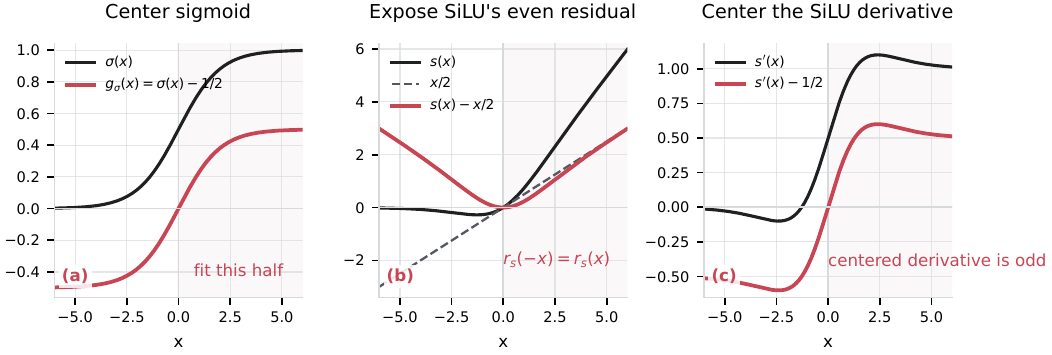}
\caption{Analytic reduction before fitting. The lightly shaded positive
half-domain is the only side that needs a polynomial. (a) Centering sigmoid
exposes an odd residual, so the negative half is recovered by a sign-bit
operation. (b) Subtracting $x/2$ from SiLU exposes an even residual despite
SiLU itself being asymmetric. (c) Centering the SiLU derivative exposes an odd
residual and enforces $s'(x)+s'(-x)=1$ by construction.}
\label{fig:symmetry-reduction}
\end{figure}

\Cref{fig:symmetry-reduction} shows why these transformations reduce more than
the sample domain. They remove coefficients that would otherwise relearn an
exact offset, linear term, or reflected half of the function. The resulting
runtime reconstruction uses absolute value, sign manipulation, and at most one
exact affine term around a short half-domain Horner chain.

\subsection{Constrained fitting and rounded-program replay}

Classical minimax construction, target-format rounding, and finite-precision
program search answer different questions. Sollya supplies the real-arithmetic
minimax control \cite{sollya}; RLIBM shows why rounding intervals belong in the
specification \cite{lim2021rlibm32}; and AutoNumerics-Zero demonstrates direct
search over executable low-precision programs \cite{real2023autonumerics}.
Our workflow combines these ideas under a fixed instruction budget.

The fitting procedure searches two boundaries. $L_i$ is the interval used to
construct the polynomial, while $L_c$ is the runtime clamp. Separating them is
useful because the least-squares endpoint and the best post-quantization tail
transition need not coincide. For each degree and $L_i$, we solve in 64-bit
floating point (FP64) a dual endpoint-constrained least-squares problem
\begin{equation}
    P(0)=f(0), \qquad P(L_i)=f(L_i).
    \label{eq:endpoint-constraints}
\end{equation}
Eliminating the first two coefficients gives the numerically convenient basis
\begin{equation}
\begin{split}
P(x)={}&f(0)+x\frac{f(L_i)-f(0)}{L_i} \\
 &+\sum_{j=2}^{D}a_j\left(x^j-xL_i^{j-1}\right),
\end{split}
\label{eq:constrained-basis}
\end{equation}
whose free basis terms vanish at both endpoints. The remaining coefficients
are solved from 2,000 Chebyshev nodes. Odd, centered-odd, and even
transformations are applied before this solve, so every candidate preserves
the corresponding global identity by construction rather than approximately.

\begin{figure}[H]
\centering
\includegraphics[
  width=\textwidth,
  alt={Three panels showing endpoint-constrained basis functions, a search over fit and clamp intervals, and error curves for exact and rounded program evaluations.}
]{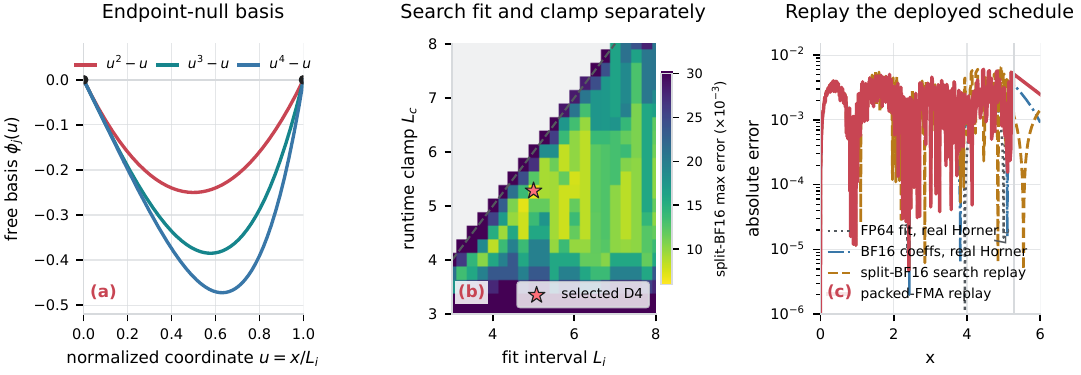}
\caption{Constrained fitting and target-program replay, illustrated with the
selected D4 centered-sigmoid candidate. (a) In normalized coordinates, every
free basis term $u^j-u$ vanishes at both endpoints, so optimization cannot
violate the two fixed values. (b) Fit interval $L_i$ and runtime clamp $L_c$
are searched independently; gray cells violate the $L_c\leq L_i+1$ search
constraint and the star marks the selected $(5.0,5.28125)$ configuration.
(c) Absolute-error curves separate the FP64 fit, BF16 coefficient rounding,
the fitter's split multiply/add replay, and a packed-FMA-style replay that
rounds once per multiply-add. The execution schedule changes both error
magnitude and where the worst error occurs.}
\label{fig:constrained-fit-replay}
\end{figure}

The three panels in \cref{fig:constrained-fit-replay} correspond to three
design choices: enforce global structure algebraically, search the active
interval jointly with the tail transition, and score the program that executes
rather than only its real-arithmetic polynomial. The BF16 search applies these
choices as follows:
\begin{enumerate}[leftmargin=1.4em]
    \item enumerate degrees D3--D6 and the function-specific $L_i,L_c$ ranges
    at 0.25-unit spacing;
    \item solve \cref{eq:constrained-basis} in FP64 on the reduced domain;
    \item round every coefficient to BF16;
    \item evaluate the configured Horner chain with target-BF16 rounding at
    each multiply and add;
    \item apply the same absolute-value reduction, clamp, sign restoration,
    and offset as the CUDA implementation;
    \item score maximum absolute error on 8,000 points over $[0,12]$, rejecting
    $L_c>L_i+1$ and any shape or boundary violation;
    \item retain distribution-weighted alternatives only after the uniform
    error and monotonicity gates pass.
\end{enumerate}
The search emulator rounds multiplication and addition as separate BF16
operations, whereas the deployed \texttt{\_\_hfma2} or PTX instruction rounds
the fused result once. Final selection replays the deployed packed-FMA order
on device.
The last panel in \cref{fig:constrained-fit-replay} also explains why a
coefficient set can improve after BF16 rounding
in real Horner evaluation yet regress once intermediate round points are
introduced. Final acceptance therefore uses the deployed packed-FMA order.

\subsection{Forward and backward programs}

We reduce forward and backward separately. Centered sigmoid is odd and its
derivative is even. Tanh is odd and its derivative is even. Subtracting $x/2$
from SiLU or exact GELU produces an even residual; subtracting $1/2$ from either
derivative produces an odd residual. We impose these identities as algebraic
fitting constraints.

For SiLU,
\begin{equation}
    s'(x)=\sigma(x)+x\sigma'(x), \qquad
    s'(x)+s'(-x)=1,
    \label{eq:silu-grad-symmetry}
\end{equation}
so $s'(x)-\tfrac12$ is odd even though the forward function is asymmetric.
Exact GELU obeys the same derivative identity. Our direct backward fit targets
this centered odd residual and preserves the complementary-gradient
relationship.

We use three backward constructions. Algebraic identities express a derivative
through a function value:
\begin{align}
    \sigma'(x) &= \sigma(x)(1-\sigma(x)), \\
    \tanh'(x) &= 1-\tanh^2(x).
\end{align}
For softcap, the scale cancels:
\begin{equation}
    \frac{\mathrm{d}}{\mathrm{d}x}
    \left[\alpha\tanh(x/\alpha)\right]
    =1-\tanh^2(x/\alpha).
    \label{eq:softcap-grad}
\end{equation}
The value can be saved from forward or recomputed in backward. B1 and B4
recompute a D3 sigmoid approximation $p(x)$, form
$\widetilde{\silu}(x)=xp(x)$, and evaluate its backward rule as
$p(x)+xp(x)(1-p(x))$. A differentiated-polynomial backward instead derives its
coefficients from the selected forward polynomial; B2 differentiates its D4
tanh program and evaluates the resulting D3 program at the recomputed score.
A direct fit to $f'$ controls gradient error independently and can exploit the
simpler backward parity. This direct program supplies a surrogate gradient
rather than the analytic derivative of the selected forward polynomial. The
accepted B3 design couples a direct D3 sigmoid value with a separately targeted,
factored D4 derivative surrogate. We measure gradient error and training loss
for the deployed construction in each case.

\subsection{Packed CUDA and inline PTX programs}

We focus on 16-bit floating-point execution because NVIDIA exposes packed
two-lane arithmetic for both IEEE binary16 (FP16) and BF16. The isolated
function sweep uses FP16: each 32-bit \texttt{\_\_half2} operand holds two
values, and \texttt{\_\_hfma2} advances two Horner chains with one packed FMA.
The B1--B4 integrations use the corresponding BF16 path, with
\texttt{\_\_nv\_bfloat162} and, for B3, explicit
\texttt{fma.rn.bf16x2} PTX.

The implementation removes scalar work from the instruction schedule in three
steps. First, B1, B2, and B4 process two BF16 values per arithmetic instruction.
On aligned workloads divisible by eight and above the vectorization threshold,
one thread loads a 16-byte \texttt{int4} chunk, reinterprets it as four
\texttt{\_\_nv\_bfloat162} pairs, and evaluates each Horner stage with
\texttt{\_\_hfma2}. Second, the fused SwiGLU program combines activation and
gating in forward and computes both projected-input gradients in one backward
kernel. These choices amortize address calculation and avoid materializing an
intermediate activation tensor.

Third, B3 places the evaluator directly inside the FA4 instruction schedule.
A separate device routine would introduce an interface boundary and hide the
register-level schedule from the surrounding CuTe kernel, so B3 uses one
\texttt{llvm.inline\_asm} block. PTX instruction names and lane semantics
follow NVIDIA's PTX instruction set architecture (ISA) \cite{nvidia2026ptxisa}.

The B3 forward block is a branch-free, two-lane BF16 evaluator. One
\texttt{cvt.rn.bf16x2.f32} packs two FP32 scores into a 32-bit register;
\texttt{min.bf16x2} and \texttt{max.bf16x2} clamp both lanes. A packed
\texttt{set.ge.u32.bf16x2} produces the piecewise mask, and
\texttt{lop3.b32} with truth table \texttt{0xca} selects embedded BF16
coefficient bit patterns independently in each lane. The mask makes region
selection branch-free, and immediate coefficient bits remove coefficient
loads. Three \texttt{fma.rn.bf16x2} instructions then evaluate the D3 Horner
chain. The coefficients are fitted
for sequence length 4096 and pre-scaled by $1/n$, so the evaluator does not
materialize $-\log n$ or apply a final normalization multiply. The complete
forward block contains 17 PTX instructions, including three packed FMAs, and
is reproduced in \cref{lst:b3-direct-d3-ptx}.

Backward reuses the forward conversion, clamp, region mask, and Horner value.
We express the factored piecewise-D4 surrogate as $P(x)(a+bx)$ and
return it from the same inline block as the D3 probability. The derivative adds
one packed FMA and one packed multiply. The coupled value/derivative block
therefore contains 22 PTX instructions and four packed FMAs, compared with 17
instructions for forward plus 22 for a standalone D4 derivative. The complete
shared schedule appears in \cref{lst:b3-coupled-d3-d4-ptx}.

\section{Experimental Setup}

All performance measurements use NVIDIA GB200. Every operator and model setting
outside the named substitution is held fixed.

\subsection{Function error and coefficient control}

The target-precision plots evaluate BF16-rounded coefficients at
BF16-representable inputs and replay the runtime Horner order. The reference
value is computed in FP32. We also compare the constrained coefficients with
Sollya's \texttt{fpminimax} \cite{sollya}. For each row, the control uses the
same degree, sparse monomials, clamp, and algebraic reconstruction. Sollya
constructs an absolute minimax polynomial with each coefficient constrained to
8-bit floating-point precision; the parsed coefficients are then cast to BF16.
The comparison in \cref{tab:function-summary} evaluates both coefficient sets
in real arithmetic on 20,001 uniformly spaced points over $[-L_c,L_c]$. This
coefficient control omits intermediate rounding and is separate from the
target-precision replay.

\subsection{Isolated function timing}

The standalone size sweep compares packed FP16 CUDA polynomial kernels with
the corresponding native PyTorch paths on GB200. Each path receives
1,000 warm-up calls. We then alternate native-first and polynomial-first order
over nine rounds of 1,000 CUDA-event-timed calls and report the ratio of the
median per-round times. The retained endpoints contain 4,194,304 elements,
whose forward and backward working sets fit in L2 cache, and 268,435,456
elements, which stream through HBM. Forward work is excluded from every
backward timing. Native sigmoid and tanh backward consume precomputed forward
outputs and apply $g\,y(1-y)$ and $g(1-y^2)$, respectively. Their polynomial
comparators consume $x$ and evaluate separately fitted D4 derivative programs.
Those two rows therefore compare direct derivative fits with native algebraic
backward, rather than comparing SFU instructions with FMAs. This sweep
exercises the FP16 \texttt{\_\_half2} implementation; the B1--B4 component and
model measurements use BF16.

\subsection{Component and training-step timing}

Speedup is
\begin{equation}
    S=\frac{T_{\native}}{T_{\poly}},
\end{equation}
so values above one favor the polynomial. The retained component benchmark
warms up for 25 iterations, then alternates native-first and polynomial-first
order over five rounds of 100 CUDA-event-timed iterations. We report the ratio
of the median per-round mean times. B1 evaluates 58,720,256 BF16 activation
elements. B2 and B3 use causal grouped-query attention (GQA) with batch 1,
sequence length 4096, 32 query heads, 8 key/value heads, and head dimension 128.
B4 evaluates 24,576 rows with hidden size 1280.

For B1--B3 full-model timing, CUDA events bracket model-plus-loss forward,
loss backward, and the complete GPU optimizer step. Each Llama~3 8B run
performs 100 optimizer steps with compilation enabled; medians use steps
20--100. Native and polynomial runs use seed 1234, batch size 4, sequence
length 4096, and the same fixed sequence of tokenized examples. Token
throughput is the training harness's wall-clock tokens per second.

B4 uses the complete DeepSeek-V3 27A4B architecture on four GB200s with expert
parallelism (EP) of four, sequence length 4096, local batch one, and global
batch four. Here 27A4B denotes approximately 27 billion total parameters and 4
billion active parameters per token. The same 100-step protocol and steps
20--100 aggregation apply. Only the 29 routed grouped-expert SwiGLU modules
change; routers, dense MLPs, and shared MLPs remain native.

\subsection{Same-checkpoint evaluation}

The auxiliary open-weight study evaluates degree-4 substitutions, which are
separate from the D3 B1 and B4 programs. Native, constrained-fit, and
same-form Sollya-fit variants run in the same process over identical requests.
The Qwen3 and Kimi checkpoint labels use A3B for approximately 3 billion
parameters active per token.
The suite records Massive Multitask Language Understanding (MMLU; 5-shot)
\cite{hendrycks2021mmlu}, HellaSwag (10-shot) \cite{zellers2019hellaswag}, the
Challenge set of the AI2 Reasoning Challenge (ARC-Challenge; 25-shot)
\cite{clark2018arc}, WinoGrande (5-shot) \cite{sakaguchi2020winogrande},
Physical Interaction: Question Answering (PIQA; 0-shot) \cite{bisk2020piqa},
Grade School Math 8K (GSM8K; 5-shot) \cite{cobbe2021gsm8k}, TruthfulQA-MC2
(0-shot) \cite{lin2022truthfulqa}, and WikiText-2 word perplexity (0-shot)
\cite{merity2017wikitext}. The reported task mean is the unweighted mean of the
seven accuracy tasks. The historical campaign launch source sets synthetic
throughput to batch 1 and sequence length 2048 for Qwen2.5; Qwen3, GLM-4.7
Flash, GPT-OSS, and Kimi use sequence length 512. Its defaults request 20
prefill measurements after five warm-ups and ten 64-token decode measurements
after two warm-ups. Results are single-run point estimates.
The retained GPT-OSS helpers use clamps of 5.25 for the constrained fit and
5.28125 for the Sollya fit.

\subsection{Paired pre-training}

Each production comparison contains one native and one polynomial trajectory.
All runs use BF16 on 32 GB200 GPUs, sequence length 4096, and 2,000 learning-rate
warm-up steps. B1--B3 use 32-way fully sharded data parallelism (FSDP) and
global batch 32; B4 uses global batch 128 and EP=8. Run fragments are ordered
and joined by token count.

The long-run metric is sampled training loss. We smooth each arm causally in
token space with an exponentially weighted mean (EWM) whose half-life is one
billion tokens, then interpolate the smoothed curves onto a shared grid spaced
every 50 million tokens through the last token count observed by both arms.
Logged training loss is the sole long-run loss metric.

\section{Results}

\subsection{Function-level approximation}

How does maximum absolute error after BF16 coefficient rounding compare with
same-form Sollya minimax controls? \Cref{tab:function-summary} evaluates both
coefficient sets with the same real-arithmetic replay.

\begin{table}[H]
\centering
\reporttablefont
\caption{Maximum absolute error ($\times10^{-3}$) for representative
constrained fits and same-form Sollya \texttt{fpminimax} controls. Both use
BF16-rounded coefficients under the same real-arithmetic replay; intermediate
target-precision rounding is evaluated in
\cref{app:extended-numerics}.}
\label{tab:function-summary}
\begin{tabularx}{\textwidth}{@{}llXrr@{}}
\toprule
& & & \multicolumn{2}{c}{\textbf{Maximum absolute error}} \\
\cmidrule(lr){4-5}
\textbf{Direction} & \textbf{Function} & \textbf{Reduced program} &
\shortstack{\textbf{Endpoint-constrained}\\\textbf{least squares}} &
\shortstack{\textbf{Sollya}\\\textbf{minimax}} \\
\midrule
Forward & Sigmoid & D3, $\tfrac12+$ odd & 5.44 & 5.47 \\
Forward & Tanh    & D4, odd & 9.61 & 7.22 \\
Forward & SiLU    & D3, $x/2+$ even & 21.74 & 19.95 \\
Forward & GELU    & D5, $x/2+$ even & 1.06 & 0.50 \\
\midrule
Backward & Sigmoid & D4, even & 4.46 & 1.97 \\
Backward & Tanh    & D4, even & 19.98 & 7.81 \\
Backward & SiLU    & D3, $\tfrac12+$ odd & 16.85 & 12.15 \\
Backward & GELU    & D5, $\tfrac12+$ odd & 8.58 & 2.36 \\
\bottomrule
\end{tabularx}
\end{table}

The constrained D3 sigmoid fit has maximum absolute error
$5.44\times10^{-3}$, compared with $5.47\times10^{-3}$ for its Sollya control.
Sollya has lower error in the other seven rows. The target-precision sweeps
measure intermediate BF16 rounding in runtime order. GELU is an auxiliary
function-level result outside the B1--B4 integrations.

The extended sweeps in
\cref{fig:forward-accuracy,fig:backward-accuracy} replay target-precision
coefficients in the runtime evaluation order. Their lower panels use
logarithmic absolute-error axes. In each of the six sweeps, at least one
higher-degree program has greater maximum error than a lower-degree program
after BF16 replay.

The backward plots include cases where a low-error value fit produces a larger
derivative error after coefficient quantization. B3 therefore pairs its D3
forward fit with a separately fitted D4 derivative. The following measurements
move from standalone functions to their consuming components.

\subsubsection{Isolated FP16 function speed}

How does the packed polynomial path compare with the native PyTorch path
before integration, and how does that comparison change with the memory
regime? \Cref{tab:function-speed} reports the cache-resident and HBM-resident
endpoints of the standalone FP16 size sweep.

\begin{table}[H]
\centering
\reporttablefont
\caption{Isolated FP16 function speed on GB200. Speedup is native PyTorch
time divided by packed polynomial CUDA time, rounded to three decimals. The two
columns report 4,194,304 elements (L2-cache resident) and 268,435,456 elements
(HBM resident). Backward polynomial rows are direct derivative fits evaluated
from $x$; native sigmoid and tanh backward use precomputed forward outputs.}
\label{tab:function-speed}
\begin{tabularx}{\textwidth}{@{}llYrr@{}}
\toprule
\textbf{Direction} & \textbf{Function} & \textbf{Polynomial program} &
\shortstack{\textbf{4M elements}\\\textbf{(L2 cache)}} &
\shortstack{\textbf{268M elements}\\\textbf{(HBM)}} \\
\midrule
Forward & Sigmoid & D3 polynomial & 1.702x & 1.249x \\
Forward & Tanh    & D4 polynomial & 1.187x & 1.143x \\
Forward & SiLU    & D3 polynomial & 1.536x & 1.267x \\
Forward & GELU    & D5 polynomial & 1.673x & 1.702x \\
\midrule
Backward & Sigmoid & D4 direct derivative & 2.135x & 1.004x \\
Backward & Tanh    & D4 direct derivative & 2.184x & 1.003x \\
Backward & SiLU    & D3 direct derivative & 2.192x & 1.119x \\
Backward & GELU    & D5 direct derivative & 2.193x & 1.615x \\
\bottomrule
\end{tabularx}
\end{table}

The cache-resident endpoint ranges from 1.187x to 2.193x. At the HBM-resident
endpoint, sigmoid and tanh backward measure 1.004x and 1.003x, while GELU
backward measures 1.615x. The sigmoid and tanh backward paths each read two
FP16 arrays and write one; their median times differ by less than 0.4\% in HBM.
Seven of the eight rows have a smaller speedup at the HBM-resident endpoint;
GELU forward measures 1.673x in L2 and 1.702x in HBM.

\subsection{Integration into LLM components}

Which function-level programs reduce component runtime after integration?
\Cref{tab:integration-summary} reports matched component measurements before
complete-model timing. B1 is a unary SiLU probe, B4 isolates the routed-expert
activation, and B2 and B3 time complete causal attention.

\begin{table}[H]
\centering
\scriptsize
\caption{B1--B4 component performance. Speedup is native time divided by
polynomial time; ratios are rounded to three decimals. Each ratio uses the
medians of five alternating, 100-iteration round means. B1 evaluates
58,720,256 values; B2 and B3 use causal grouped-query attention (GQA) with batch
1, sequence length 4096, 32 query heads, 8 key/value heads, and head dimension
128; B4 evaluates 24,576 rows with hidden size 1280. The B1 row is a unary
direct-derivative probe; the complete-step B1 measurements use the deployed
packed SwiGLU path described below.}
\label{tab:integration-summary}
\begin{tabularx}{\textwidth}{@{}lYYrr@{}}
\toprule
\textbf{Case} & \textbf{Timed component} & \textbf{Program} &
\shortstack{\textbf{Forward}\\\textbf{speedup}} &
\shortstack{\textbf{Backward}\\\textbf{speedup}} \\
\midrule
B1 & One-kernel SiLU activation & D3 direct-fit probe & 1.055x & 1.007x \\
B2 & Complete causal GQA attention & D4 tanh softcap & 2.053x &
1.046x \\
B3 & Complete causal GQA attention & D3/D4 sigmoid & 1.074x &
1.001x \\
B4 & One-kernel expert activation & D3 SwiGLU & 1.003x & 1.005x \\
\bottomrule
\end{tabularx}
\end{table}

\subsubsection{B1: dense SiLU}

The unary D3 SiLU probe measures 1.055x forward speedup and 1.007x backward
speedup on 58.7M BF16 values; its backward is the standalone direct D3
derivative fit. The deployed B1 packed SwiGLU backward instead recomputes the
D3 sigmoid approximation from the gate input and forms the SiLU derivative
algebraically. The complete-step measurements below exercise that deployed
path.

\subsubsection{B2: tanh softcap}

B2 replaces tanh in \cref{eq:softcap} inside FA4. Forward uses packed D4.
Backward differentiates the selected D4 coefficients and evaluates the
resulting D3 polynomial at the recomputed, scaled attention score. Complete
causal attention measures 2.053x forward speedup and 1.046x backward speedup.

\subsubsection{B3: sigmoid attention}

The direct sigmoid program accelerates complete attention forward. For sequence
length $n$, attention score $s_{ij}$ is mapped to probability $P_{ij}$, which
weights the value matrix $V$ to produce output $O$:
\begin{equation}
    P_{ij}=\sigma(s_{ij}-\log n), \qquad O=PV.
    \label{eq:sigmoid-attention}
\end{equation}
At sequence length 4096, the evaluator computes this normalized sigmoid with
D3 forward and a coupled, factored D4 derivative surrogate; a runtime guard
enforces that fitted length. Complete-attention forward measures 1.074x and
backward measures 1.001x.

\subsubsection{B4: routed expert SwiGLU}

B4 applies D3 SiLU inside routed-expert SwiGLU. The router implementation
remains native; only the activation inside the routed-expert operation changes.
The fused expert operation is
\begin{equation}
    Y=\widetilde{\silu}(XW_g)\odot(XW_u).
\end{equation}
Its backward recomputes the D3 sigmoid approximation from the gate input,
forms SiLU and its derivative algebraically, and computes gradients for both
projected inputs in one pass. The one-kernel activation benchmark measures
1.003x in forward and 1.005x in backward.

\paragraph{Auxiliary integrations.}
Further simplification of FA4's fractional exponential measures 1.898x for its
isolated fractional core and 1.0008x at the model step; complete fits and
timings appear in \cref{app:exp2-negative}. Appendix
\ref{app:rope-negative} describes the packed sine/cosine implementation and its
integration with rotary position embedding (RoPE).

\FloatBarrier
\subsection{Model-scale results}

\subsubsection{Complete-model timing}

How do phase speedups and training-throughput ratios change at complete-model
scale?
\Cref{tab:model-timing-summary} separates forward, backward, GPU-step, and
wall-clock throughput ratios.

\begin{table}[H]
\centering
\scriptsize
\caption{Complete-model performance. B1--B3 use matched 100-step Llama~3 8B
probes on one GB200. B4 uses the complete DeepSeek-V3 27A4B architecture on
four GB200s. Timing speedup is native time divided by polynomial time;
training-throughput ratio is polynomial tokens/s divided by native tokens/s.
Ratios use medians over steps 20--100, are rounded to three decimals, and favor
the polynomial above one.}
\label{tab:model-timing-summary}
\begin{tabularx}{\textwidth}{@{}lYrrrr@{}}
\toprule
\textbf{Case} & \textbf{Model / intervention} &
\shortstack{\textbf{Forward}\\\textbf{speedup}} &
\shortstack{\textbf{Backward}\\\textbf{speedup}} &
\shortstack{\textbf{GPU-step}\\\textbf{speedup}} &
\shortstack{\textbf{Training-throughput}\\\textbf{ratio}} \\
\midrule
B1 & Llama~3 8B, dense D3 SiLU & 1.005x & 1.038x &
1.027x & 1.027x \\
B2 & Llama~3 8B, D4 tanh softcap & 1.107x & 1.004x &
1.029x & 1.029x \\
B3 & Llama~3 8B, D3/D4 sigmoid attention & 1.011x & 0.997x &
1.003x & 1.002x \\
B4 & DeepSeek-V3 27A4B, expert D3 SwiGLU & 1.087x &
1.077x & 1.080x & 1.080x \\
\bottomrule
\end{tabularx}
\end{table}

B1 and B2 retain complete-step gains in the Llama~3 probes. B3's
attention-forward change is smaller at the complete-model step. B4 improves
both full-model phases in the 27A4B probe.

\subsubsection{Same-checkpoint open-weight evaluation}

How do fixed-checkpoint task accuracy, word perplexity, and inference throughput
change under the degree-4 substitutions? \Cref{tab:eval} compares constrained
and same-form Sollya coefficients through the same degree and integration site.

\begin{table}[H]
\centering
\scriptsize
\caption{Same-checkpoint degree-4 ablations. ``Endpoint LS'' is this paper's
endpoint-constrained least-squares fit; ``Sollya minimax'' is the same-form
control. Mean accuracy is the unweighted mean of seven tasks; its change is
variant minus native and is rounded to 0.01 percentage points. WikiText-2 word perplexity is
native$\rightarrow$variant and is rounded to three decimals. Throughput ratios
are variant divided by native, are rounded to three decimals, and favor the
polynomial above one.}
\label{tab:eval}
\resizebox{\textwidth}{!}{%
\begin{tabular}{@{}lllrrrr@{}}
\toprule
\textbf{Model} & \textbf{Operator} &
\shortstack{\textbf{Fit}\\\textbf{method}} &
\shortstack{\textbf{Mean accuracy}\\\textbf{change}\\\textbf{(percentage points)}} &
\shortstack{\textbf{WikiText-2}\\\textbf{word perplexity}} &
\shortstack{\textbf{Prefill}\\\textbf{throughput ratio}} &
\shortstack{\textbf{Decode}\\\textbf{throughput ratio}} \\
\midrule
Qwen2.5 7B & Dense SwiGLU & Endpoint LS & -0.51 &
$8.734\!\rightarrow\!197.677$ & 1.035x & 1.013x \\
 & & Sollya minimax & -39.80 &
$8.734\!\rightarrow\!298{,}700.735$ & 1.036x & 1.015x \\
\addlinespace
Qwen3 30B-A3B & Expert SwiGLU & Endpoint LS & -0.10 &
$7.648\!\rightarrow\!7.655$ & 1.056x & 1.025x \\
 & & Sollya minimax & +0.04 &
$7.648\!\rightarrow\!7.651$ & 1.051x & 1.025x \\
\addlinespace
GLM-4.7 Flash & Expert SwiGLU + router sigmoid & Endpoint LS & -0.30 &
$11.406\!\rightarrow\!11.400$ & 0.969x & 1.004x \\
 & & Sollya minimax & -0.18 &
$11.406\!\rightarrow\!11.614$ & 0.966x & 1.003x \\
\addlinespace
GPT-OSS 120B & Expert SiLU epilogue & Endpoint LS & +0.41 &
$414.947\!\rightarrow\!407.545$ & 1.001x & 0.993x \\
 & & Sollya minimax & -0.08 &
$414.947\!\rightarrow\!411.659$ & 0.999x & 0.993x \\
\addlinespace
Kimi Linear 48B-A3B & Expert SwiGLU + router sigmoid & Endpoint LS & +0.04 &
$7.102\!\rightarrow\!7.101$ & 1.084x & 1.013x \\
 & & Sollya minimax & -0.09 &
$7.102\!\rightarrow\!7.100$ & 1.084x & 1.009x \\
\bottomrule
\end{tabular}%
}
\end{table}

Qwen3 and Kimi improve prefill by 5.1--5.6\% and 8.4\%, respectively; their
task-mean changes span $-0.10$ to $+0.04$ percentage points and their WikiText-2
perplexity changes by at most 0.0071. Qwen2.5 improves prefill by 3.5--3.6\%
while its word perplexity rises from 8.734 to 197.677 and 298,700.735. GLM
prefill measures 0.969x and 0.966x, and GPT-OSS measures 1.001x and 0.999x.
The Qwen2.5 rows also show that similar kernel throughput across coefficient
sources can accompany sharply different model outputs.

\subsubsection{Paired pre-training to 100 billion tokens}

\begin{figure}[H]
    \centering
    \includegraphics[
      width=\linewidth,
      alt={Four panels compare native and polynomial smoothed pre-training loss through approximately 100 billion tokens for cases B1 through B4, with late-stage insets.}
    ]{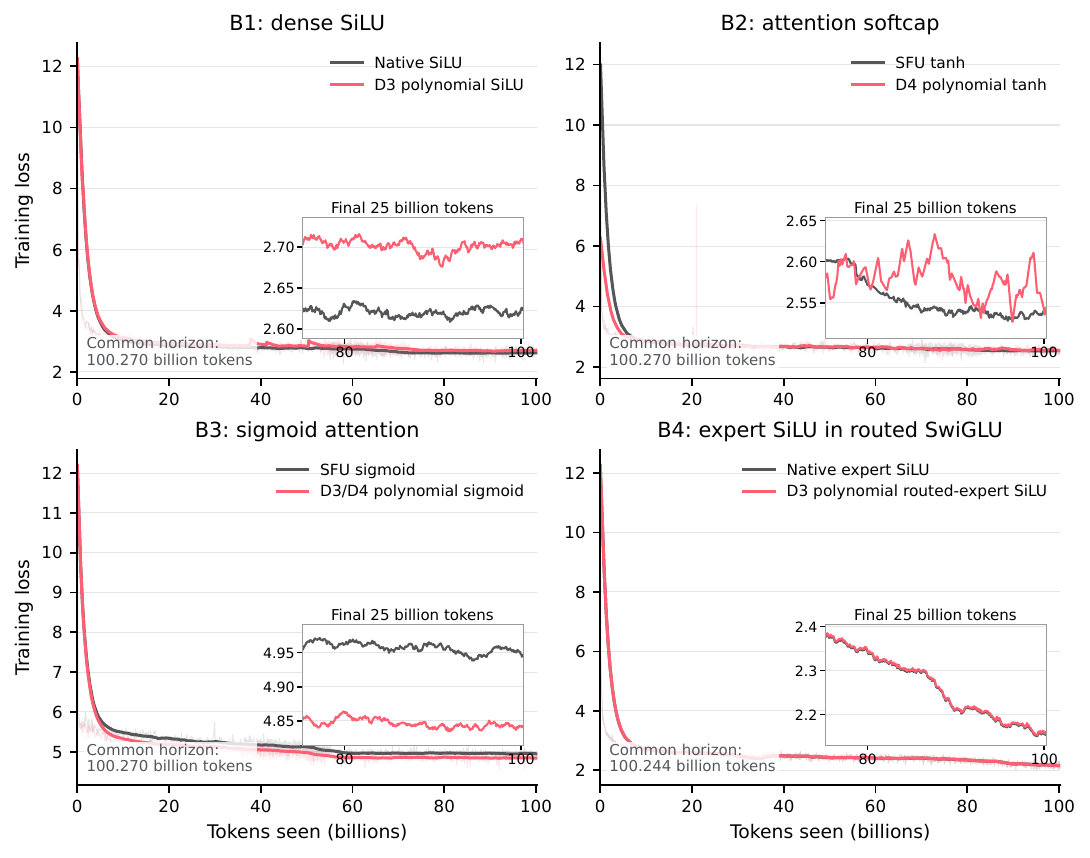}
    \caption{Paired B1--B4 pre-training loss. Gray is the native/SFU control
    and coral is the polynomial candidate. Thin lines are uniformly thinned
    sampled training loss; thick lines are token-domain exponentially weighted
    means (EWMs) with a half-life of one billion tokens, interpolated after
    smoothing onto a shared grid with 50 million tokens between points. Each
    panel ends at the common observed horizon. Insets show the final 25 billion
    tokens.}
    \label{fig:paired-production-loss}
\end{figure}

How do the training-loss trajectories change when the polynomial is used
throughout pre-training? \Cref{tab:pretraining-summary} reports the final EWM
value at the last token count shared by each native/polynomial pair, and
\cref{fig:paired-production-loss} shows the complete trajectories.

\begin{table}[H]
\centering
\reporttablefont
\caption{Paired pre-training at the common observed horizon. Loss is the
exponentially weighted mean (EWM) of sampled training loss with a half-life of
one billion tokens. Training-loss change is polynomial minus native. Losses and
changes are rounded to four decimals. Each row is one paired trajectory.}
\label{tab:pretraining-summary}
\begin{tabularx}{\textwidth}{@{}lYXrrr@{}}
\toprule
\textbf{Case} & \textbf{Intervention} &
\shortstack{\textbf{Shared horizon}\\\textbf{(billion tokens)}} &
\shortstack{\textbf{Native EWM}\\\textbf{training loss}} &
\shortstack{\textbf{Polynomial EWM}\\\textbf{training loss}} &
\shortstack{\textbf{Training-loss}\\\textbf{change}} \\
\midrule
B1 & Dense D3 SiLU & 100.2701 & 2.6264 & 2.7050 & $+0.0786$ \\
B2 & D4 tanh softcap & 100.2701 & 2.5415 & 2.5439 & $+0.0025$ \\
B3 & D3/D4 sigmoid attention & 100.2701 & 4.9471 & 4.8397 & $-0.1073$ \\
B4 & Routed-expert D3 SwiGLU & 100.2439 & 2.1532 & 2.1560 & $+0.0028$ \\
\bottomrule
\end{tabularx}
\end{table}

At the common horizons near 100 billion tokens, B2 and B4 finish within 0.003
final EWM training-loss units of their native pairs. B1 finishes 0.0786 higher
and B3 finishes 0.1073 lower.

\FloatBarrier
\section{Discussion and Conclusion}

Across B1--B4, the phase-level measurements differ by integration site. Dense
SiLU measures its larger gain in backward, tanh softcap in forward, and
routed-expert SwiGLU in both. Sigmoid attention improves its forward kernel
while the complete model step changes much less. The auxiliary fractional
exponential measures 1.898x in isolation and 1.0008x at the model step after
range reduction, data movement, and fusion are included.

The measurements cover GB200, BF16/FP16 programs, four principal integration
sites, model-dependent same-checkpoint substitutions, and paired pre-training
trajectories measured with training loss. Function,
component, and model measurements evaluate symmetry, coefficient rounding,
backward rules, packed instruction schedules, and the consuming operator as one
integrated kernel program. Additional engineering observations are collected
in \cref{app:lessons-and-boundaries}, and the auxiliary integrations appear in
\cref{app:negative-results}.

As matrix, special-function, and memory pipelines change across hardware
generations, this workflow locates transcendental work that remains exposed in
modern generative modelling.

\section*{Acknowledgements}

The author thanks Tom Cashman, Paul Balança, Douglas Orr, Carlo Luschi, and
Andrew Fitzgibbon for discussions and feedback around this work.

\section*{Generative-AI Disclosure}

Generative AI tools assisted with code development, experiment orchestration,
analysis, figure generation, and drafting and editing this report. The author
reviewed and verified the code, experiments, numerical results, figures,
citations, and final text and takes responsibility for the content.

\begingroup
\scriptsize
\bibliographystyle{plainnat}
\bibliography{main}
\endgroup

\clearpage
\appendix

\section{Extended Numerical Results}
\label{app:extended-numerics}

The complete D3--D6 sweeps below replay BF16-rounded coefficients in the
runtime evaluation order at BF16-representable inputs. Forward and derivative
fits are evaluated separately.

\begin{figure}[H]
\centering
\begin{subfigure}[b]{0.49\textwidth}
    \includegraphics[width=\linewidth,alt={Sigmoid function value and absolute error over BF16-representable inputs for polynomial degrees D3 through D6.}]{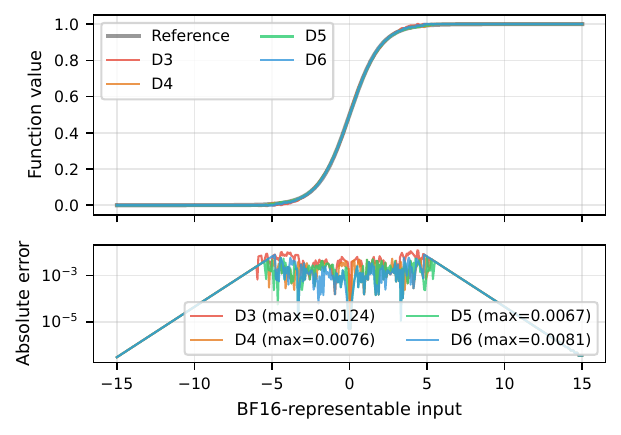}
    \caption{Sigmoid forward.}
\end{subfigure}
\hfill
\begin{subfigure}[b]{0.49\textwidth}
    \includegraphics[width=\linewidth,alt={Tanh function value and absolute error over BF16-representable inputs for polynomial degrees D3 through D6.}]{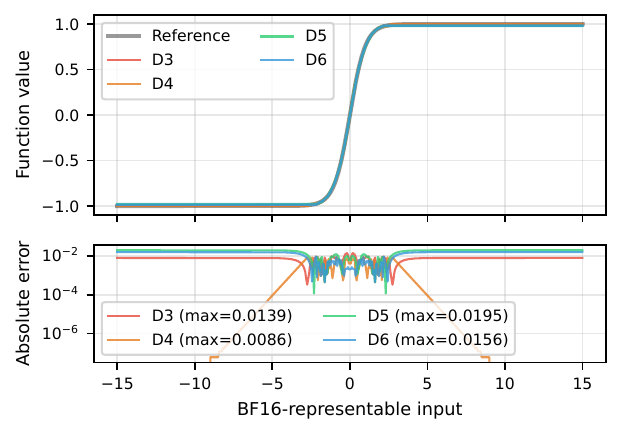}
    \caption{Tanh forward.}
\end{subfigure}
\par\medskip
\begin{subfigure}[b]{0.49\textwidth}
    \includegraphics[width=\linewidth,alt={SiLU function value and absolute error over BF16-representable inputs for polynomial degrees D3 through D6.}]{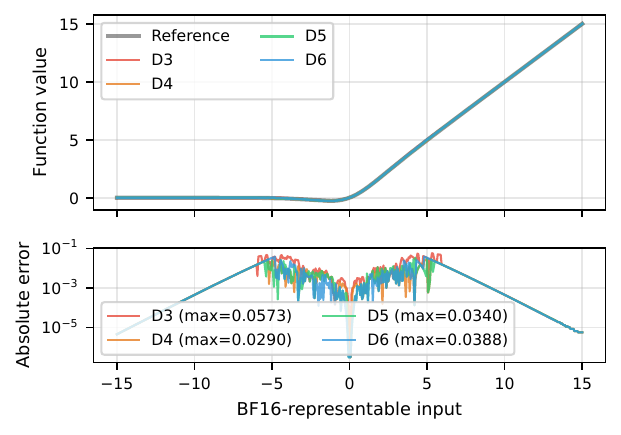}
    \caption{SiLU forward.}
\end{subfigure}
\caption{Forward numerical accuracy for D3--D6 BF16 target-precision programs.
Absolute error is measured against the FP32 reference at BF16-representable inputs.}
\label{fig:forward-accuracy}
\end{figure}

\begin{figure}[H]
\centering
\begin{subfigure}[b]{0.49\textwidth}
    \includegraphics[width=\linewidth,alt={Sigmoid derivative value and absolute error over BF16-representable inputs for polynomial degrees D3 through D6.}]{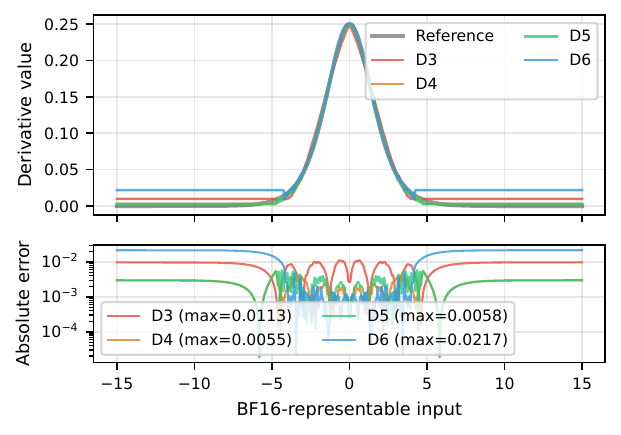}
    \caption{Sigmoid derivative.}
\end{subfigure}
\hfill
\begin{subfigure}[b]{0.49\textwidth}
    \includegraphics[width=\linewidth,alt={Tanh derivative value and absolute error over BF16-representable inputs for polynomial degrees D3 through D6.}]{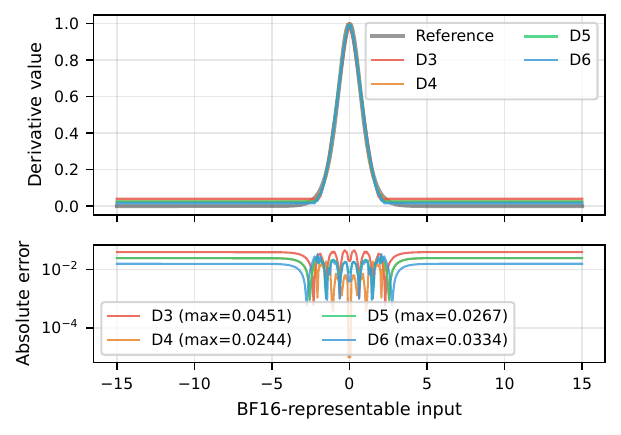}
    \caption{Tanh derivative.}
\end{subfigure}
\par\medskip
\begin{subfigure}[b]{0.49\textwidth}
    \includegraphics[width=\linewidth,alt={SiLU derivative value and absolute error over BF16-representable inputs for polynomial degrees D3 through D6.}]{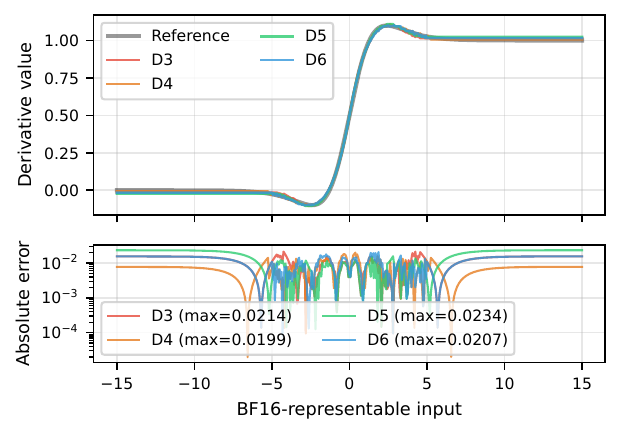}
    \caption{SiLU derivative.}
\end{subfigure}
\caption{Backward numerical accuracy for D3--D6 BF16 target-precision programs.
Absolute error is measured against the FP32 reference at BF16-representable inputs.}
\label{fig:backward-accuracy}
\end{figure}

\FloatBarrier

\section{Auxiliary Integration Results}
\label{app:negative-results}

Two auxiliary experiments pair inner-arithmetic measurements with the
surrounding FA4 and rotary-position-embedding operations.

\subsection{Further \texttt{exp2} simplification in FlashAttention-4}
\label{app:exp2-negative}

FA4 decomposes $x$ into an integer and a fractional component. Its software
path writes
\begin{equation}
    x=n+f,\qquad n=\lfloor x\rfloor,\quad f\in[0,1),\qquad
    2^x=2^n 2^f,
    \label{eq:exp2-reduction}
\end{equation}
then approximates only $2^f$. The integer scale $2^n$ is reconstructed from
the exponent bits. All comparisons below retain FA4's FP32 lower clamp, range
reduction, exponent reconstruction, masking rules, and mixed SFU/software
routing. Only the fractional evaluator changes within the existing attention
contract.

\paragraph{Fits and invariants.}
Online softmax repeatedly rescales its running row state when the row maximum
changes. A fractional approximation must therefore satisfy $p(0)=1$,
$p(1)=2$, and continuity between adjacent exponent intervals. These conditions
leave only $p_1(f)=1+f$ for a single line. For a two-piece piecewise-linear
(PWL2) approximation, we use the selection-free hinge form
\begin{equation}
    p_{\mathrm{PWL2}}(f)=1+af+2(1-a)\max(f-\tfrac12,0),
    \qquad a=0.7980513010,
    \label{eq:exp2-pwl2-safe}
\end{equation}
and for the quadratic control we use
\begin{equation}
    p_{\mathrm{D2}}(f)=1+af+(1-a)f^2,
    \qquad a=0.6602339719.
    \label{eq:exp2-d2-safe}
\end{equation}
The parameters in \cref{eq:exp2-pwl2-safe,eq:exp2-d2-safe} minimize sampled
maximum relative error on a $1{,}000{,}001$-point grid. We also solve the same
one-parameter families by least squares to expose the objective trade-off;
least squares gives $a=0.7852775264$ for PWL2 and $a=0.6563658611$ for D2.
For comparison, FA4's existing D3 is
\begin{equation}
    p_{\mathrm{D3}}(f)=1+0.6951461434f+0.2275643945f^2
    +0.0771190897f^3.
    \label{eq:exp2-d3-fa4}
\end{equation}
\Cref{tab:exp2-fits} compares both objectives with FA4's existing
Sollya/minimax D3 polynomial. The D2 sampled-minimax fit is more accurate than
PWL2 despite having the same two-FMA Horner depth because it needs no hinge.

\begin{table}[H]
\centering
\reporttablefont
\caption{Fractional $2^f$ fits on $f\in[0,1]$. Errors are measured in FP64 on
a dense grid before coefficient quantization. ``Core'' excludes common range
reduction and exponent reconstruction. D1, PWL2, and D2 preserve both
endpoints exactly.}
\label{tab:exp2-fits}
\begin{tabularx}{\textwidth}{@{}llXrr@{}}
\toprule
\textbf{Family} & \textbf{Objective} & \textbf{Fractional core} &
\shortstack{\textbf{Maximum relative}\\\textbf{error}} &
\shortstack{\textbf{Root-mean-square}\\\textbf{relative error}} \\
\midrule
D1 & Endpoint constrained & 1 packed FMA & $6.148\times10^{-2}$ & $4.466\times10^{-2}$ \\
PWL2 & Least squares & 2 packed FMAs + hinge & $1.526\times10^{-2}$ & $6.754\times10^{-3}$ \\
PWL2 & Sampled minimax & 2 packed FMAs + hinge & $1.074\times10^{-2}$ & $7.174\times10^{-3}$ \\
D2 & Least squares & 2 packed FMAs & $3.204\times10^{-3}$ & $1.987\times10^{-3}$ \\
D2 & Sampled minimax & 2 packed FMAs & $2.679\times10^{-3}$ & $1.921\times10^{-3}$ \\
FA4 D3 & Sollya/minimax & 3 packed FMAs; $p(1)=1.99983$ & $8.763\times10^{-5}$ & $6.029\times10^{-5}$ \\
\bottomrule
\end{tabularx}
\end{table}

The packed FP16 PWL2 coefficients are $a=0.7978515625$ and
$2(1-a)=0.40380859375$. They preserve both endpoints exactly after rounding
and have maximum relative error $1.132\times10^{-2}$. The handwritten core is
11 inline PTX instructions, including two \texttt{fma.rn.f16x2}
instructions. The D1 core needs only one packed FMA, but its fit error is much
larger and it is no faster than D2 in causal FA4. The BF16 PWL2 core is also
slower and less accurate than FP16.

For online softmax, endpoint constraints are part of the approximation
contract: $p(0)=1$ preserves the identity update and $p(1)=2$ preserves an
integer exponent step. The accepted PWL2 comparison therefore includes only
fits that satisfy both constraints after target-precision rounding. Pointwise
error and endpoint invariants are both carried through the recurrent state
transition.

\paragraph{Isolated evaluator.}
\Cref{tab:exp2-evidence} separates fractional-core timing from the complete
exponential. The PWL2 fractional timing uses the reported coefficient set; the
constrained fit has the same instruction shape and latency. Repeated
register-resident arithmetic exposes the shorter fractional core, while
L2-sized and HBM-sized working sets and the shared FP32 reduction and
reconstruction change the measured balance. PWL2 replaces the saved FMA with
hinge subtraction and maximum instructions.

\begin{table}[t]
\centering
\reporttablefont
\caption{The exp2 performance ladder on GB200. Entries are speedups over the
matching native/all-SFU path. Full-exp rows include unchanged FP32 reduction
and reconstruction; FA4 rows use identical causal routing. Values are rounded
to three decimals, except the 8B optimizer-step probes, which are rounded to
four decimals.}
\label{tab:exp2-evidence}
\begin{tabularx}{\textwidth}{@{}Xrrr@{}}
\toprule
\textbf{Measurement scope} & \textbf{FA4 D3} & \textbf{PWL2 FP16} &
\textbf{D2 FP32} \\
\midrule
Fractional core, repeated compute & -- & 1.898x & -- \\
Full $2^x$, repeated compute & 0.884x & 0.863x & -- \\
Full $2^x$, 4 MiB input + 4 MiB output (L2-sized) & 0.995x & 1.002x & -- \\
Full $2^x$, 256 MiB input + 256 MiB output (HBM-sized) & 0.995x & 0.990x & -- \\
Causal FA4 forward & 1.012x & 1.010x & 1.012x \\
Causal FA4 forward + backward & 1.001x & 1.005x & 1.006x \\
80-step 8B optimizer, opposite-order geometric mean & -- & 1.0008x & 1.0030x \\
\bottomrule
\end{tabularx}
\end{table}

For the FA4 accuracy comparison, tensor-relative $L_2$ error is
$\lVert y_{\mathrm{variant}}-y_{\mathrm{SFU}}\rVert_2 /
\max(\lVert y_{\mathrm{SFU}}\rVert_2,10^{-30})$. We report the maximum over the
attention output and the query, key, and value input gradients.

\begin{table}[H]
\centering
\reporttablefont
\caption{Maximum tensor-relative $L_2$ error against all-SFU FA4 on GB200.}
\label{tab:exp2-accuracy}
\begin{tabularx}{\textwidth}{@{}Xrrr@{}}
\toprule
\textbf{Measured quantity} & \textbf{FA4 D3} & \textbf{PWL2 FP16} &
\textbf{D2 FP32} \\
\midrule
Maximum over output and input-gradient tensors & $8.641\times10^{-4}$ &
$3.458\times10^{-3}$ & $2.019\times10^{-3}$ \\
\bottomrule
\end{tabularx}
\end{table}

\paragraph{FA4 forward and backward.}
The integrated target is FA4, which deliberately
interleaves a minority of software evaluations with SFU work to balance two
execution pipelines. The integrated rows of \cref{tab:exp2-evidence} compare
D3, PWL2, and D2 under one routing contract. The causal Llama-8B GQA geometry is
$B=1,S=4096,H_q=32,H_{kv}=8,D=128$, where $B$ is batch size, $S$ is sequence
length, $H_q$ is the number of query heads, $H_{kv}$ is the number of key/value
heads, and $D$ is head dimension. Timings use a requested 1300 MHz
streaming-multiprocessor clock (1305 MHz observed), and speedups are medians of
15 paired per-round
ratios with 100 iterations per round. The matched rows route 33.3\% of
eligible forward and 12.5\% of backward lane pairs through software. The
all-SFU reference measures 0.2749~ms forward and 0.9174~ms forward plus
backward; matched D3, PWL2, and D2 measure 0.2715/0.9169,
0.2723/0.9127, and 0.2714/0.9113~ms, respectively.

The compared paths differ in four relevant ways. FA4's D3 core uses three
packed FP32 FMAs and runs concurrently with SFU work rather than replacing a
serial native exponential. PWL2 replaces one FMA with a hinge subtraction and
maximum, and its FP16 path adds conversions. Every software variant retains
clamp, floor emulation, fractional extraction, and exponent-bit reconstruction.
Only a minority of eligible lane pairs use software: final fragments and causal
masked tiles remain on SFU, while query--key (QK) and probability--value (PV)
matrix multiplication, online reductions, and synchronization are unchanged.

\paragraph{8B model probes.}
Opposite-order 80-step Llama-8B probes use the same sequence length and clock.
PWL2 forward-plus-backward measures 1.0025x when the SFU control runs first and
0.9990x in reverse order, for a 1.0008x geometric mean. D2 measures 0.9904x
and 1.0158x. The opposed orders span a wider range than their geometric means
in \cref{tab:exp2-evidence}. Final candidate-minus-control loss deltas range
from -0.323 to -0.050 after 80 optimizer steps.

\paragraph{Exposure analysis for further exp2 degree reduction.}
The bottleneck can be quantified with Amdahl's law. If the isolated fractional
core accelerates by $s=1.898$ but matched FA4 forward accelerates by only
$S=1.010$, the effective serial share exposed to that optimization is
\begin{equation}
    p=\frac{1-S^{-1}}{1-s^{-1}}\approx 0.021.
    \label{eq:exp2-amdahl}
\end{equation}
The measured saving is 2.6~$\mu$s in FA4 forward and 4.7~$\mu$s in
forward plus backward. Even multiplying the latter by all 32 Llama layers
gives approximately 0.15~ms against a roughly 275~ms optimizer step, about
0.05\% of step time, consistent with the complete-step timing.

Blackwell Ultra doubles \texttt{MUFU.EX2} throughput for key attention instructions and
advertises up to 2x attention-layer compute over Blackwell
\cite{nvidia2026blackwellultrasoftmax,aubrey2025blackwellultra}.

\FloatBarrier

\subsection{Paired sine/cosine and fused RoPE}
\label{app:rope-negative}

Rotary position embedding (RoPE) supplies a paired-function experiment because
sine is odd, cosine is even, and both share one periodic reduction. For angular
frequency $\omega_i$, the host stores the unsigned Q0.32 turn increment
$\Delta_i=\operatorname{round}(\omega_i2^{32}/2\pi)$ in a 32-bit word. At
position $p$, the kernel forms $(p\Delta_i)\bmod 2^{32}$, reduces it to a signed
Q15 half-turn coordinate, and evaluates a packed FP16 D5 sine and D6 cosine with
shared squared input. Each thread processes four frequencies as two packed
pairs. The fitting, numerical-check, repeated-evaluator, and fused-RoPE drivers
are under \path{src/sfu_repro/rope/} in the standalone reproduction repository.
The retained timing summaries do not identify a final quantitative comparison,
so this appendix reports the implementation and integration boundary.

\FloatBarrier

The fused experiment reads contiguous FP16 Q in $[B,S,H_q,D]$ and K in
$[B,S,H_{kv},D]$, where $B$ is batch size, $S$ is sequence length, $H_q$ is the
number of query heads, $H_{kv}$ is the number of key/value heads, and $D$ is
head dimension. It computes four sine/cosine pairs once per thread, reuses them
across the corresponding heads, applies the packed complex rotation, and writes
the transposed layouts $[B,H_q,S,D]$ and $[B,H_{kv},S,D]$. Cached-table,
on-the-fly native-SFU, and on-the-fly polynomial variants share the same
rotation and memory instructions.

Inspection with NVIDIA's binary utilities can distinguish the polynomial
instruction path from SFU, FP32-FMA, or packed-BF16 fallbacks
\cite{nvidia2026binaryutilities}. Query/key movement, packed complex rotation,
transpose, and store remain shared across variants.

\FloatBarrier

\clearpage
\section{Packed PTX Listings}
\label{app:packed-ptx}

The following listings are emitted by the B3 inline-PTX generator in
\path{flash-attention/flash_attn/cute/handwritten_spline_ptx.py}. These are
generator-emitted source-PTX listings; reported counts cover executable
instructions and exclude register declarations.

\begin{lstlisting}[
  style=reportcode,
  language=PTX,
  label={lst:b3-direct-d3-ptx},
  caption={Source PTX for the fixed-length ($n=4096$) B3 direct-D3
  FlashSigmoid forward evaluator. Two FP32 scores are rounded into one BF16x2
  register; packed comparison and \texttt{lop3.b32} select one of two
  coefficient rows, and three packed FMAs evaluate the cubic. Hex immediates
  contain duplicated BF16 values after $1/4096$ coefficient scaling.}
]
{
    .reg .b32 %fa4_input, %fa4_mask_lo, %fa4_mask_hi,
        %fa4_mask, %fa4_clamped, %fa4_value, %fa4_active;
    cvt.rn.bf16x2.f32 %fa4_input, $2, $1;
    mov.b32 %fa4_mask_lo, 0x3fa03fa0;
    set.ge.u32.bf16x2 %fa4_mask, %fa4_input, %fa4_mask_lo;
    mov.b32 %fa4_mask_lo, 0x40c040c0;
    min.bf16x2 %fa4_clamped, %fa4_input, %fa4_mask_lo;
    mov.b32 %fa4_mask_lo, 0xc0c0c0c0;
    max.bf16x2 %fa4_clamped, %fa4_clamped, %fa4_mask_lo;
    lop3.b32 %fa4_value, %fa4_mask,
        0x3a093a09, 0x37e837e8, 0xca;
    lop3.b32 %fa4_mask_hi, %fa4_mask,
        0xbb11bb11, 0x39093909, 0xca;
    fma.rn.bf16x2 %fa4_value, %fa4_clamped,
        %fa4_value, %fa4_mask_hi;
    lop3.b32 %fa4_mask_hi, %fa4_mask,
        0x3b8c3b8c, 0x39873987, 0xca;
    fma.rn.bf16x2 %fa4_value, %fa4_clamped,
        %fa4_value, %fa4_mask_hi;
    lop3.b32 %fa4_mask_hi, %fa4_mask,
        0xbb09bb09, 0x39803980, 0xca;
    fma.rn.bf16x2 %fa4_value, %fa4_clamped,
        %fa4_value, %fa4_mask_hi;
    mov.b32 %fa4_mask_lo, 0;
    max.bf16x2 %fa4_value, %fa4_value, %fa4_mask_lo;
    mov.b32 $0, %fa4_value;
}
\end{lstlisting}

\begin{lstlisting}[
  style=reportcode,
  language=PTX,
  label={lst:b3-coupled-d3-d4-ptx},
  caption={Source PTX for the coupled B3 direct-D3 probability and
  factored-D4 derivative evaluator. Packing, region selection, clamping, and
  the complete three-FMA D3 chain are shared. The derivative tail selects a
  region-specific linear factor, evaluates it with one packed FMA, and
  multiplies it by the D3 value.}
]
{
    .reg .b32 %fa4_input, %fa4_mask_lo, %fa4_mask_hi,
        %fa4_mask, %fa4_clamped, %fa4_value, %fa4_grad;
    cvt.rn.bf16x2.f32 %fa4_input, $3, $2;
    mov.b32 %fa4_mask_lo, 0x3fa03fa0;
    set.ge.u32.bf16x2 %fa4_mask, %fa4_input, %fa4_mask_lo;
    mov.b32 %fa4_mask_lo, 0x40c040c0;
    min.bf16x2 %fa4_clamped, %fa4_input, %fa4_mask_lo;
    mov.b32 %fa4_mask_lo, 0xc0c0c0c0;
    max.bf16x2 %fa4_clamped, %fa4_clamped, %fa4_mask_lo;
    lop3.b32 %fa4_value, %fa4_mask,
        0x3a093a09, 0x37e837e8, 0xca;
    lop3.b32 %fa4_mask_hi, %fa4_mask,
        0xbb11bb11, 0x39093909, 0xca;
    fma.rn.bf16x2 %fa4_value, %fa4_clamped,
        %fa4_value, %fa4_mask_hi;
    lop3.b32 %fa4_mask_hi, %fa4_mask,
        0x3b8c3b8c, 0x39873987, 0xca;
    fma.rn.bf16x2 %fa4_value, %fa4_clamped,
        %fa4_value, %fa4_mask_hi;
    lop3.b32 %fa4_mask_hi, %fa4_mask,
        0xbb09bb09, 0x39803980, 0xca;
    fma.rn.bf16x2 %fa4_value, %fa4_clamped,
        %fa4_value, %fa4_mask_hi;
    mov.b32 %fa4_mask_lo, 0;
    max.bf16x2 %fa4_value, %fa4_value, %fa4_mask_lo;
    lop3.b32 %fa4_mask_hi, %fa4_mask,
        0xbb38bb38, 0xbc10bc10, 0xca;
    lop3.b32 %fa4_mask_lo, %fa4_mask,
        0x3f813f81, 0x3f803f80, 0xca;
    fma.rn.bf16x2 %fa4_grad, %fa4_clamped,
        %fa4_mask_hi, %fa4_mask_lo;
    mul.rn.bf16x2 %fa4_grad, %fa4_value, %fa4_grad;
    mov.b32 $0, %fa4_value;
    mov.b32 $1, %fa4_grad;
}
\end{lstlisting}

\FloatBarrier

\section{Lessons and Boundaries}
\label{app:lessons-and-boundaries}

\begin{enumerate}[leftmargin=1.4em]
    \item \textbf{Generated instruction schedules determine latency.}
    High-level CuTe expressions sometimes lower to long instruction sequences;
    the workflow inspects the generated PTX.
    \item \textbf{Backward has its own error profile.} Direct backward fitting
    is used when the differentiated forward approximation exceeds the training
    error budget.
    \item \textbf{Algorithmic invariants accompany pointwise error.}
    Online-softmax approximations preserve interval endpoints alongside the
    scalar error bound.
    \item \textbf{Integration determines exposure.} The fractional
    exponential core measures 1.898x before its surrounding FA4 work and
    1.0008x at the model step. B1 measures 1.005x in full forward and 1.038x
    in backward, producing a 1.027x complete-step result.
\end{enumerate}

\section{Code and Reproducibility}
\label{app:reproducibility}

\begingroup
\scriptsize
The report builds with \texttt{make -C report}; the artifact map and regeneration
commands are in \path{report/README.md}. The standalone reproduction repository is
\url{https://github.com/MrHuff/fast-polynomial-transcendentals}. It contains the
paper-specific integration layer without the original model-training framework,
plus the fitting, packed-kernel, open-weight, and materialized-evidence paths.
Filenames below without a directory are under \path{report/data/}.

\begin{description}[leftmargin=1.55in,style=nextline]
    \item[Function error]
    \path{autonumerics_zero/cuda_benchmarks/analysis_results/sollya_device_bf16.json}
    contains the constrained and same-form Sollya comparison.

    \item[Isolated function timing]
    \path{isolated_function_speedups_fp16.csv} contains the FP16 endpoint
    ratios in \cref{tab:function-speed}, and
    \path{isolated_function_speedups_fp16.provenance.json} records the
    benchmark protocol and manuscript lineage. Raw alternating-round timings
    and environment metadata are in
    \path{isolated_function_speedups_fp16_gb200.json}. The runnable driver is
    \path{autonumerics_zero/experiments/benchmark_report_function_speed.py}.

    \item[Integrated timing]
    \path{b1_b4_100_iteration_phase_probes_gb200.json},
    \path{b1_b4_full_model_100step_gb200.json}, and
    \path{b4_27a4b_local_100step_gb200.json} contain the component and
    complete-model probes.

    \item[Paired pre-training]
    \path{b1_b4_paired_loss_curves.csv} contains the sampled histories used in
    \cref{fig:paired-production-loss};
    \path{b1_b4_paired_loss_curves.provenance.json} records the retained run
    fragments. Complete launch manifests for initialization and data order are
    not part of the artifact set.

    \item[Open-weight evaluation]
    \path{sollya_eval_results.csv} contains \cref{tab:eval};
    \path{open_weight_eval_protocol.json} records the source-derived launch
    configuration and runtime-attestation status.

    \item[Implementations]
    In the standalone repository, fitting code is under
    \path{autonumerics_zero/evolution/}, packed kernels are under
    \path{autonumerics_zero/spline_ops/}, portable B1--B4 plumbing is under
    \path{src/sfu_repro/}, and B3's inline-PTX source is in the pinned
    \path{flash-attention} submodule.
\end{description}

\endgroup

\end{document}